\RequirePackage{fix-cm}
\documentclass{svjour3} 
\smartqed  

\usepackage{graphicx}
\usepackage{amsmath}
\usepackage{hyperref}
\usepackage{xcolor}
\usepackage{amsmath,amssymb,amsfonts}
\usepackage{algorithmic}
\usepackage{textcomp}
\usepackage{xcolor}
\usepackage{subfig}
\usepackage{gensymb}
\usepackage[numbers]{natbib}
\usepackage{caption}

\usepackage{mwe}
\usepackage{amsfonts}
\usepackage{amssymb}
\usepackage{verbatim}
\usepackage{amsbsy}
\usepackage{siunitx}
\usepackage{tabularx, booktabs}
\usepackage{soul}
\usepackage{tikz}
\usetikzlibrary{calc}
\usepackage{microtype}

\usepackage{alphalph}

\begin{document}

\title{Double Critic Deep Reinforcement Learning for Mapless 3D Navigation of Unmanned Aerial Vehicles}

\titlerunning{Double Critic Deep-RL for Mapless 3D Navigation of Unmmaned Aerial Vehicles}        

\thanks{$^{1}$ Ricardo Bedin Grando are with the Technological Unit, Techonological University of Uruguay -- UTEC, Rivera, Uruguay.
E-mail: {\tt\small ricard.bedin@utec.edu.uy}
}
\thanks{$^{2}$Junior Costa de Jesus and Paulo L. J. Drews-Jr are with the NAUTEC, Centro de Ciências Computacionais, Univ. Fed. do Rio Grande -- FURG, RS, Brazil.
E-mail: {\tt\small paulodrews@furg.br}
}
\thanks{$^{3}$Victor Augusto Kich and Alisson Henrique Kolling are with Federal University of Santa Maria, Brazil.
E-mail: {\tt\small victorkich@yahoo.com.br}
}

\author{Ricardo Bedin Grando$^{1}$, Junior Costa de Jesus$^{2}$, Victor Augusto Kich$^{3}$, Alisson Henrique Kolling$^{3}$, Paulo Lilles Jorge Drews-Jr$^{2}$
}

\authorrunning{Ricardo Bedin Grando} 

\institute{
Ricardo Bedin Grando \at 
Technological University of Uruguay\newline
Rivera, Uruguay\newline
\email{ricardo.bedin@utec.edu.uy}\and
Junior Costa de Jesus \and Paulo Lilles Jorge Drews-Jr \at
Federal University of Rio Grande\newline
Rio Grande, RS, Brazil\newline
\email{dranaju@gmail.com, paulodrews@furg.br}\and
Victor Augusto Kich \and Alisson Henrique Kolling\at
Federal University of Santa Maria\newline
Santa Maria, RS, Brazil\newline
\email{victorkich@yahoo.com.br, alikolling@gmail.com}
}

\date{Received: date / Accepted: date}

\maketitle
\begin{abstract} 
This paper presents a novel deep reinforcement learning-based system for 3D mapless navigation for Unmanned Aerial Vehicles (UAVs). Instead of using a image-based sensing approach, we propose a simple learning system that uses only a few sparse range data from a distance sensor to train a learning agent. We based our approaches on two state-of-art double critic Deep-RL models: Twin Delayed Deep Deterministic Policy Gradient (TD3) and Soft Actor-Critic (SAC). We show that our two approaches manage to outperform an approach based on the Deep Deterministic Policy Gradient (DDPG) technique and the BUG2 algorithm. Also, our new Deep-RL structure based on Recurrent Neural Networks (RNNs) outperforms the current structure used to perform mapless navigation of mobile robots. Overall, we conclude that Deep-RL approaches based on double critic with Recurrent Neural Networks (RNNs) are better suited to perform mapless navigation and obstacle avoidance of UAVs.

\keywords{Twin Delayed Deep Deterministic Policy Gradients \and Soft Actor-Critic \and Deep Reinforcement Learning \and Navigation for Aerial Vehicles}
\end{abstract}

\section{Introduction} 
\label{intro}

In recent years, Deep Reinforcement Learning (Deep-RL) has been employed in a wide range of fields, achieving interesting results in many tasks for control of discrete systems \cite{dulac2015deep} and continuous systems \cite{lillicrap2015continuous}. More recently, the learning was applied in tasks in robotics as well, where it was initially used to handle tasks in stable and observable environments \cite{duan2016benchmarking}. For mobile robotics, however, the complexity increases significantly given the interactions with barriers in the physical workplace \cite{drews2016,xie2017towards}. In this context, Deep-RL ended up simplifying the problem by discretizing it \cite{zhu2017target}. Other approaches have also been exploring continuous control actions in the navigation of mobile robots and achieving interesting results \cite{tai2017virtual}, \cite{chen2017socially}, \cite{jesus2019deep}. However, the results are yet limited for autonomous Unmanned Aerial Vehicles (UAVs) \cite{sampedro2019fully,kang2019generalization}

In our previous work, we proposed the adaptation of two Deep-RL techniques for an UAV in two environments \cite{grando2020deep} and explored the concept for mobile robots \cite{de2021soft}. Based on the state-of-art structure for terrestrial mobile robots \cite{tai2017virtual}, we investigated the performance of Deep-RL for 2D mapless navigation related tasks of UAVs. However, we propose a new structure for the agents for 3D navigation and develop a new deterministic approach based on the Twin Delayed Deep Deterministic Policy Gradient (TD3) \cite{fujimoto2018addressing} extending our previous work. Instead of using the conventional structure used to perform mapless navigation of terrestrial mobile robots, we propose a new methodology with Recurrent Neural Networks (RNNs) with more sophisticated Deep-RL approaches.

\begin{figure}[!htb]
\centering
\includegraphics[width=\linewidth]{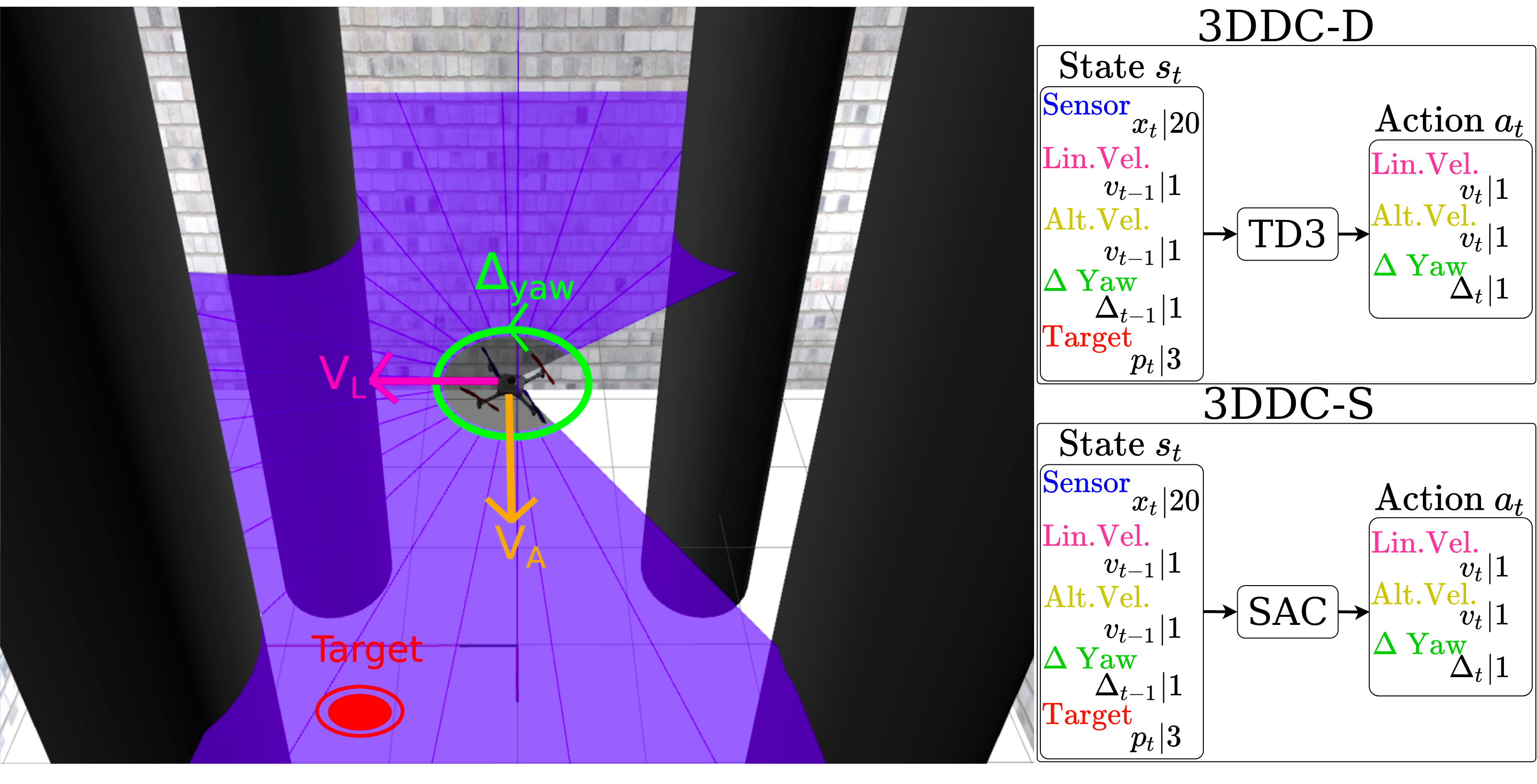}
\caption{Our second simulated environment (left) and Deep-RL structure (right).}
\label{fig:system_def}
\end{figure}

The objective of this work is to show how well Deep-RL approaches based on RNNs can perform in 3D navigation-related tasks. In this work we also include environmental difficulties, such as simulated wind, to improve the overall robustness. Our approach was done in such a way that the network used had only 26 inputs and 3 outputs, as shown in Figure \ref{fig:system_def}. As inputs, it was used readings of a distance sensor coupled to the upper part of the vehicle, the previous linear velocities, the delta yaw angle and the distance and angle of the UAV related to the target. The actions were set as the outputs of the network to send it to the vehicle to arrive at the target. We expect that the intelligent agent will not just be able to arrive at the target goal but also be able to avoid any possible collision on the way to get it. We also expect the agents to perform through the environmental difficulties, showing robustness on the adaptation. Our improved deterministic approach (3DDC-D) and our bias stochastic approach (3DDC-S) are set to be compared with our previous Deep-RL approaches and with a behavior-based approach inspired in the work of \cite{marino2016minimalistic} that can be used to perform mapless navigation of UAVs. \footnote{\scalebox{1}{Code available at \href{https://github.com/ricardoGrando/hydrone\_deep\_rl\_jint}{\textcolor{gray}{https://github.com/ricardoGrando/hydrone\_deep\_rl\_jint}}.}}\footnote{\scalebox{1}{Video available at \href{https://www.youtube.com/watch?v=Kf5T2SgyiRs}{\textcolor{gray}{https://www.youtube.com/watch?v=Kf5T2SgyiRs}}.}}

This work has the following contributions:

\begin{itemize}

\item We present a novel methodology that only uses range-data readings and the vehicle's relative localization data to successfully perform goal-oriented mapless 3D navigation and waypoint 3D mapless navigation for an UAV. 

\item We show that our new methodology based on RNNs improves the results obtained from the state-of-the-art Deep-RL for terrestrial robots \cite{tai2017virtual} and in our previous work \cite{grando2020deep}.

\item We present a novel deterministic approach based on the double critic actor-critic TD3 algorithm that outperforms the single critic DDPG algorithm used in previous works \cite{grando2020deep}, \cite{bedin2021deep}.

\item We show that our agents manage to learn environmental difficulties, such as wind, showing that learning-based approaches are more robust and adapt better than behavior-based algorithms.

\item We show that with our novel methodology the robot improves its ability to avoid collision with obstacles, outperforming our previous approaches in the avoidance capability. 

\end{itemize}


The work has the following structure: related works are addressed in the Sec. \ref{related_works}. Our novel approaches can be seen in the the Sec. \ref{methodology}. Then, in the Section \ref{experimental_setup} the tools and techniques that we used are presented. The results achieved and the comparison with out previous structures and algorithms are presented in the Sec. \ref{results}. For the last, Sec. \ref{conclusion} shows our our overall analysis of the results and the perspective for future works.

\section{Related Works}
\label{related_works}

Deep-RL surveys in robotics tasks for the navigation problem have already been performed to demonstrate how effectively we can solve this problem with learning approaches \cite{kober2013reinforcement}, \cite{kormushev2013reinforcement}, \cite{tobin2017domain}, \cite{bedin2021deep}.

Tai \emph{et al.} \cite{tai2017virtual} used 10-dimensional range findings and a goal location as inputs and continuous steering commands as outputs for a 2D mapless motion planner when applied to mobile robots. They concluded that an approach based on the DDPG method with a fully-connected ANN of three hidden layers would effectively train a 2D mapless motion planner. Furthermore, the process of completing the task of reaching a predetermined goal was completed successfully.

Zhu \emph{et al.} \cite{zhu2017target} used an actor-critic Deep-RL model to tackle the challenge of driving a mobile robot. It took the current state observation and an image of the target as inputs, producing the action in a 3D environment as output. Chen \emph{et al.} \cite{chen2017socially} discussed the issue of navigation in a pedestrian-rich environment. It proposed a Deep-RL approach to control a mobile robot that could navigate in a speed close to a human walking speed. It also performed in a scenario with multiple humans simulating a crowd. Ota \emph{et al.} \cite{ota2020efficient} developed an approch based on the SAC technique to efficiently navigate. It was also done in simulation, with the agent learning the course.

For UAVs, however, Deep-RL has been addressed fewer times. Rodriguez \emph{et al.} \cite{rodriguez2018deep} used DDPG to solve the problem of landing on a moving platform. On the Gazebo simulator, they used Deep-RL in conjunction with the RotorS framework \cite{furrer2016rotors} for aerial vehicles. Sampedro \emph{et al.} \cite{sampedro2019fully} suggested that an approach based on the DDPG technique could perform a task of Search and Rescue in an indoor environment. Sampedro \emph{et al.} \cite{sampedro2019fully} made the use of a real and simulated UAV to perform this navigation task. Kang \emph{et al.} \cite{kang2019generalization} focused on tasks realted to avoid collision with obstacles. This work also used visual information and real and simulated data.

He \emph{et al.} \cite{he2020deep} combines imitation learning and TD3 technique and applied on the challenging 3D UAV navigation problem using depth cameras and sketched in a variety of simulation environments, obtaining great results. Li \emph{et al.} \cite{li2020uav} combines Deep-RL with meta-learning and proposes a novel approach, named Meta Twin Delayed Deep Deterministic Policy Gradient (Meta-TD3), to realize the control of an UAV. Thereafter, Li \emph{et al.} compared your method with DDPG and TD3, obtaining a substantial improvement in terms of both convergence value and convergence rate.

Differently, our method addresses 3D mapless navigation and waypoint mapless navigation tasks by relying solely on laser sensor readings and the vehicle's relative localization data. Since the TD3 and SAC techniques are state-of-art for terrestrial mobile robots, we based our approach on them. We also make use of a LSTM-based structure instead of a MLP structure, showing that RNN based structures are better suited for this problem. We also differ by imposing environmental difficulties such as wind, which makes the overall results worse but more realistic.

\section{Methodology}
\label{methodology}

In this work, we propose a system for UAVs that can learn to navigate from a starting position on space to a target position and through a series of target points, building its own motion plan on environments with obstacles. It was designed in such a way that it only needs a couple of range sensing data and the vehicles' relative localization data to perform the mapless navigation and the obstacle avoidance. Its movement equation can be defined as:

\begin{equation}
\upsilon_t = f(x_t, p_t, \upsilon_{t-1}),
\end{equation}

where $x_t$ is the raw information from the sensor readings, $p_t$ the relative position and angle of the target and $\upsilon_{t-1}$ the vehicle's velocity in the last step. This model allows obtaining the actions that the robot is able to make, given its current state $s_t$. Embodied in a neural network, the expected result is an action for the current state.
This arrange for the data was explored in Tai \textit{et al} \cite{tai2017virtual}, Jesus \textit{et al} in \cite{jesus2019deep} and in our previous work \cite{grando2020deep} for 3D navigation of UAV.

\subsubsection{Networks Structure}

\begin{figure}[ht]
\centering
\includegraphics[width=0.5\textwidth]{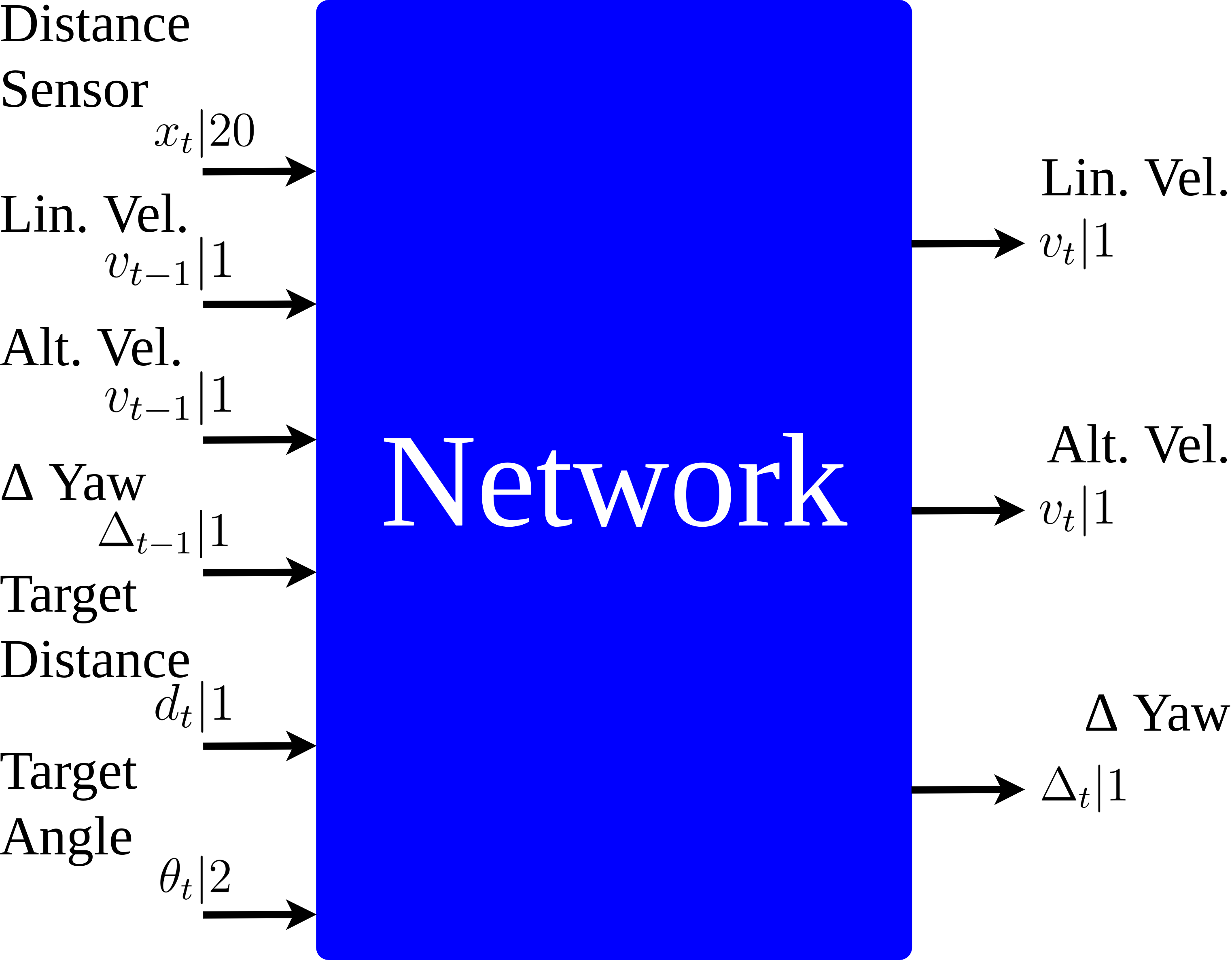}
\caption{Network Inputs and Outputs.}
\label{fig:network_strucuture}
\end{figure}

We defined a structure that has 26 inputs and 3 outputs. Among the 26 inputs, 20 represent range findings from a simulated Lidar, 3 represent the previous action and the other 3 represent information regarding the target goal. Our Lidar provides 1080 samples in a $270\degree$ range, 20 of which sampled equally spaced in $13.5\degree$. The target information used was the vehicles' relative distance to the goal and two relative angles to the target, one given by the plan $x-y$ and the second given by the plan $z-distance$. Both the relative distance and angles were used to make the agent learns towards their minimization and achieve the goal. The network's outputs provide the action for the given step, which are the linear and altitude velocities and the variation of the yaw angle. These actions are then applied to the vehicle. Figure \ref{fig:network_strucuture} shows the proposed architecture.

\begin{figure}[ht]
\centering
\includegraphics[width=0.5\textwidth]{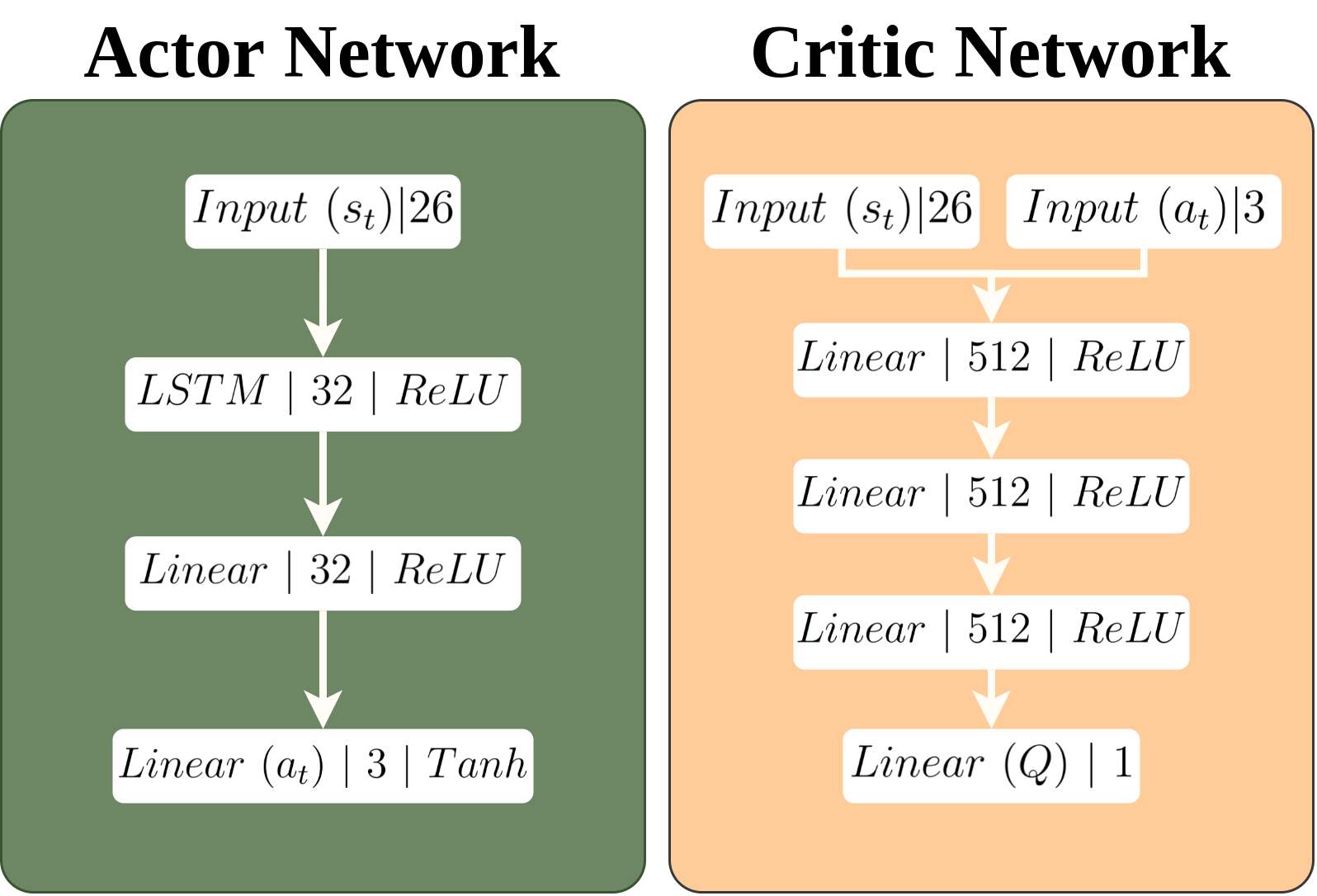}
\caption{Proposed new model architecture.}
\label{fig:new_stucture}
\end{figure}

Our newly proposed network architecture has a 32-cell LSTM layer for the actor-network activated using the ReLU function. The hyperbolic tangent function was used as the activation function in the actor's outputs. The values are then scaled between $0$ and $0.25$ $m/s$ for the linear velocity, $-0.25$ and $0.25$ $rads$ for the delta yaw and $-0.25$ and $0.25$ $m/s$ for the altitude velocity. The Q-value for the current state and action performed by the agent is given in the critic-network. We projected the Deep-RL-based agents similarly so we compare the more fairly the characteristics of each approach. For the SAC-based approach, specifically, the value network was set with the same structure as the critic network. Figure \ref{fig:new_stucture} shows our new proposed network structure while Figure \ref{fig:tai_structure} shows the previous architecture proposed by Tai \textit{et al.} \cite{tai2017virtual}, also used in our previous work \cite{grando2020deep}.

\subsubsection{Reward Function}

With the environment and the agents defined, it is feasible to train and evaluate the agents. Before training, it is necessary to create a reward and penalty system for the agent. These values given to the agents are numbers of a function model, based on empirical knowledge, created during the process of solution of the problem. 

In our previous work, we defined a system that there were three conditions. In this current work, one of the conditions regarding the distance in each step was not used. As consequence, it caused an increased training time. However, this simplified reward function presented better results at the end of training when compared with the previous one that was proposed. The reward function used in this work was the following:
\begin{equation}
r (s_t, a_t) = 
\begin{cases}
r_{arrive} \ \textrm{if} \ d_t < c_d
\\
r_{collide} \ \textrm{if}\ min_x < c_o
\end{cases}
\end{equation}

Only two rewards were given, one for making the task correctly and the second one in case of failure. The agent receives a positive $r_{arrive}$ reward of $100$ when the goal is reached in a margin of $c_d$ meters This margin, specifically, was set as $0.5m$. Also, in case of a collision with an obstacle or with the limits of the scenario, we give a negative reward $r_{collide}$ of $-10$. The collisions is verified if the readings from the distance sensor is lower than a distance $c_o$ of $0.5$ meters. This simplified rewarding system also helps to focus on the Deep-RL approaches themselves, their similarities and differences, instead of focusing on the environment.

\section{Comparative Approaches}

\begin{figure}[ht]
\centering
\includegraphics[width=0.5\textwidth]{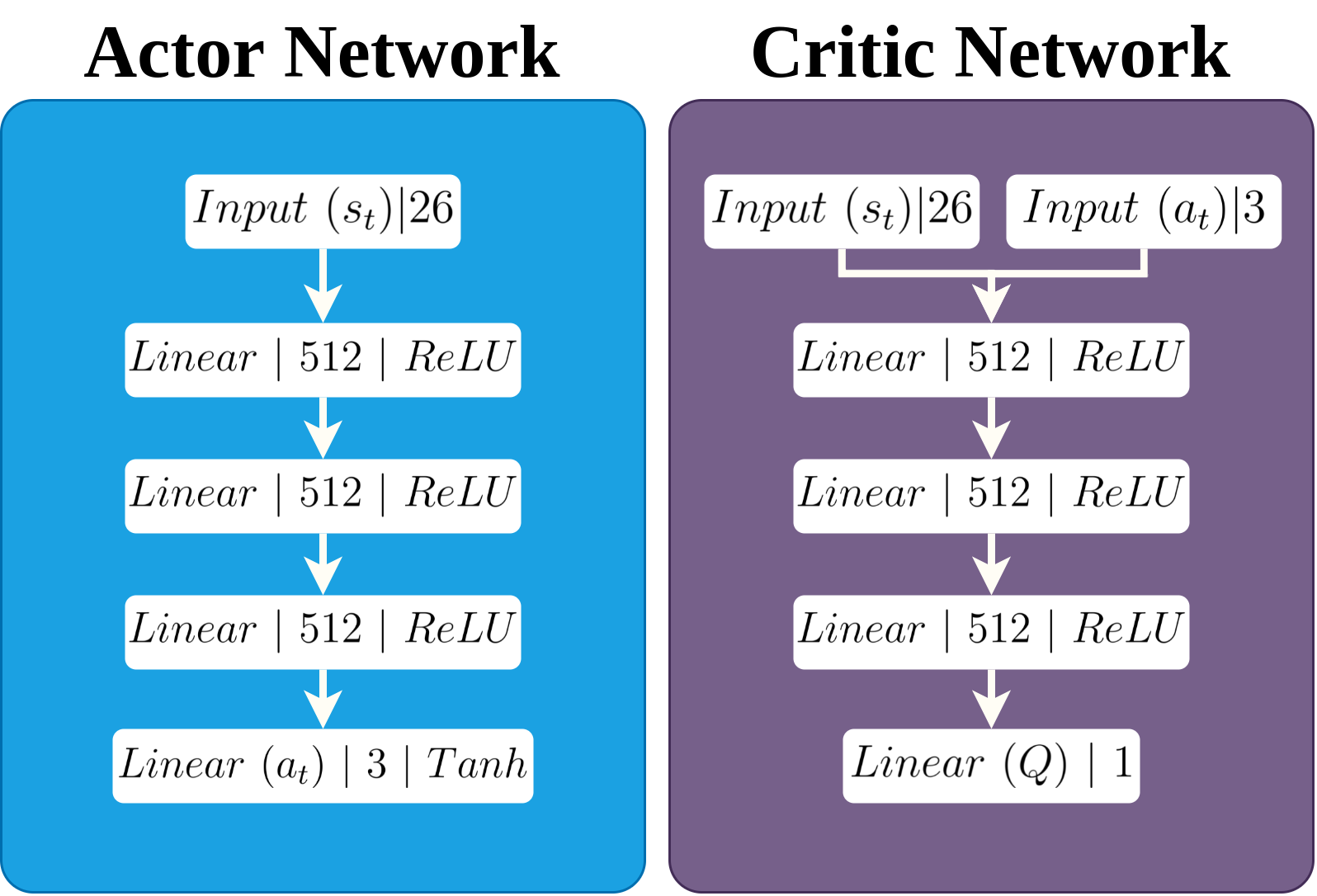}
\caption{3-Layers MLP model architecture.}
\label{fig:tai_structure}
\end{figure}

In one of our latest works \cite{bedin2021deep}, we proposed two 3D quad-rotor navigation approaches: a deterministic based on DDPG, and a stochastic based on SAC, both using MLP3. These algorithms are named 3DNDRL-D and 3DNDRL-S, respectively. These approaches have the same structure like the ones that we used in our previous work \cite{grando2020deep}, only adapted to perform the 3D mapless navigation of a hybrid vehicle instead of the 2D mapless navigation of an UAV. The networks of 3DNDRL-D and 3DNDRL-S approaches have 3 hidden layers each. The structure used was a fully-connected multi-layer perception, where each layers has 512 neurons each. the ReLU activation was used as well. Fig \ref{fig:tai_structure} summarizes this structures.

The other algorithm that we used to compare was inspired in the work of Marino \emph{et al.} \cite{marino2016minimalistic}. It uses a BUG-like algorithm to perform a 2D simulated multi-floor indoor navigation with a quad-rotor robot, equipped with a salient-cue sensor and a laser-range-finder. As well known, the BUG2 tries to minimize the distance between the robot and the goal, toward the estimated position of the goal on the horizontal plane while avoiding obstacles. So, it is possible to implement a local navigation approach for navigating in the horizontal plane.

\section{Experimental Setup}
\label{experimental_setup}

For the analysis of this work, in this section is described the setup, tools and materials that were used. 
Thus, making it clear the programming language and framework utilized for our simulation, as well as, to describe the environments purposed for the task of navigation of a hybrid UAV concept.

\subsection{Trainig Setup}
In this article, the experiments carried out were programmed using the Python programming language. The neural networks were developed with PyTorch\cite{pytorch}, which provides tensor computation with acceleration via GPU and an automatic differentiation system and has a varied range of applications such as in fields of natural language processing, image processing, object recognition, and others\cite{rukhovich2021iterdet,Tao2020HierarchicalMA,shoeybi2020megatronlm}.   
For the robot simulation, we used the Robot Operating System(ROS)  \cite{quigley2009ros} together with Gazebo. ROS is a framework for developing applications for robots with a collection of tools and libraries that provides operating system standard services, such as hardware abstraction, device low-level control and package management. For this work, it was used the Melodic version of ROS. The gazebo is an open-source 3D robotics simulator, that provides an integrated physics simulator, 3D rendering and support for the integration and simulation of sensors and controllers, has available plugins for the simulation of aerial autonomous vehicles and is fully integrated with ROS.

\subsection{Vehicle's description}

The experiments were conducted utilizing an UAV, which consists of a quad-rotor initially developed as a hybrid vehicle capable of transit through air and water, as shown in previous works\cite{drews2014,neto2015}. This hybrid UAV, improved with new models and propellers\cite{horn2019study,horn_lars2020} and with the Ardupilot and ROS-based control system concept for our developed HUAUV \cite{grando2019deep}, succeeded to win the 2019 Brazil Open Flying Robot League during the 2019 Latin America Robotics Competition\cite{grando_competi2020}. 


To simulate our vehicle, we created a 3D model and a universal robot description. The package containing the description and the Deep-RL approaches are provided in code\footnote{\scalebox{0.8}{$https://github.com/ricardoGrando/hydrone\_deep\_rl\_jint$}}. We used the RotorS framework to create the description of the vehicle using real dimensions and constants (such as mass, inertia, motor constants, etc). The visual meshes are based on the model that was developed in Solidworks software. And we used a simplified collision structure, approximated by cubes and cylinders. 


From the network, our robot receives a linear velocity, an altitude velocity and a $\Delta$yaw. To control the robot, we used the RotorS' internal geometric tracking controller. The conversion from the linear and angular velocities and the $\Delta$yaw to Cartesian velocities is needed, so a plugin was developed to convert from polar to Cartesian coordinates. 
With this structure, the agents managed to learn the vehicle's properties, suppress the environmental difficulties and perform navigation-related tasks.

\subsection{Simulated Environments}

The development of applications for robots is usually difficult to conceive and maintain given the high cost of hardware and equipment. Many applications in robotics are then developed in simulation and then deployed to real robots. Therefore, a good simulation environment is necessary to make progress in the field. Gazebo simulator \cite{koenig2004design} can be used to fulfill this need, given its extensive compatibility and realistic simulation.

\begin{figure}[!ht]
  \subfloat[First environment.\label{fig:env1}]{
	\begin{minipage}[c][0.65\width]{
	   0.5\textwidth}
	   \centering
	   \includegraphics[width=\textwidth]{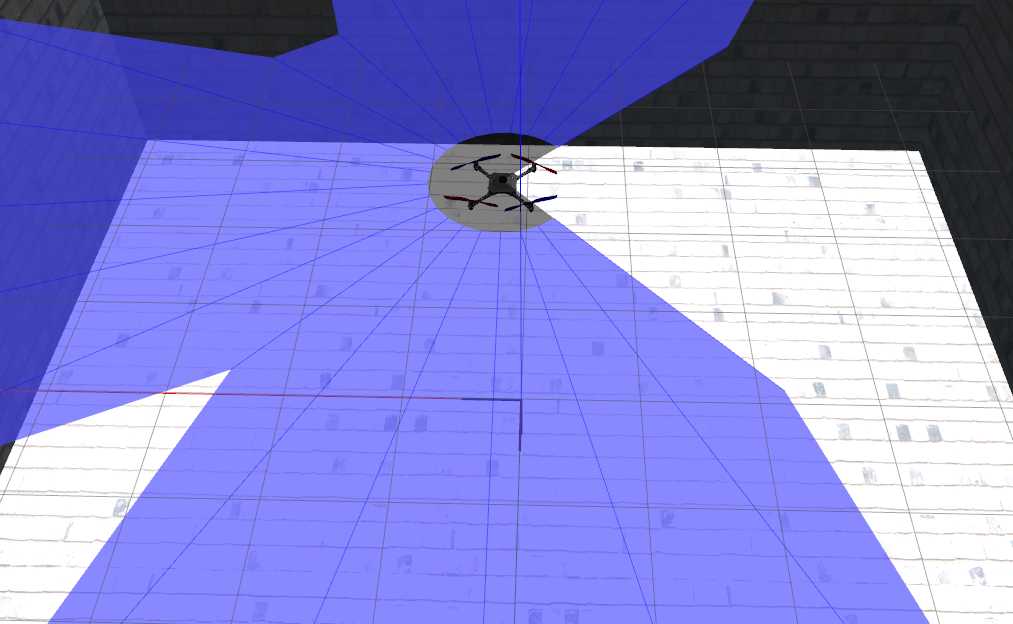}
	\end{minipage}}
 \hfill 
  \subfloat[First environment from above.\label{fig:env1_upper}]{
	\begin{minipage}[c][0.65\width]{
	   0.5\textwidth}
	   \centering
	   \includegraphics[width=0.62\textwidth]{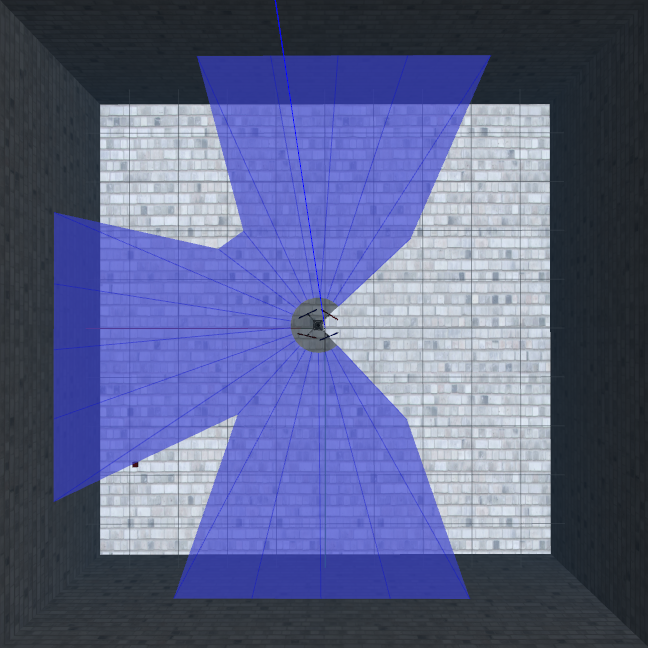}
	\end{minipage}}
 \hfill
  \subfloat[Second environment\label{fig:env2}]{
	\begin{minipage}[c][0.65\width]{
	   0.5\textwidth}
	   \centering
	   \includegraphics[width=\textwidth]{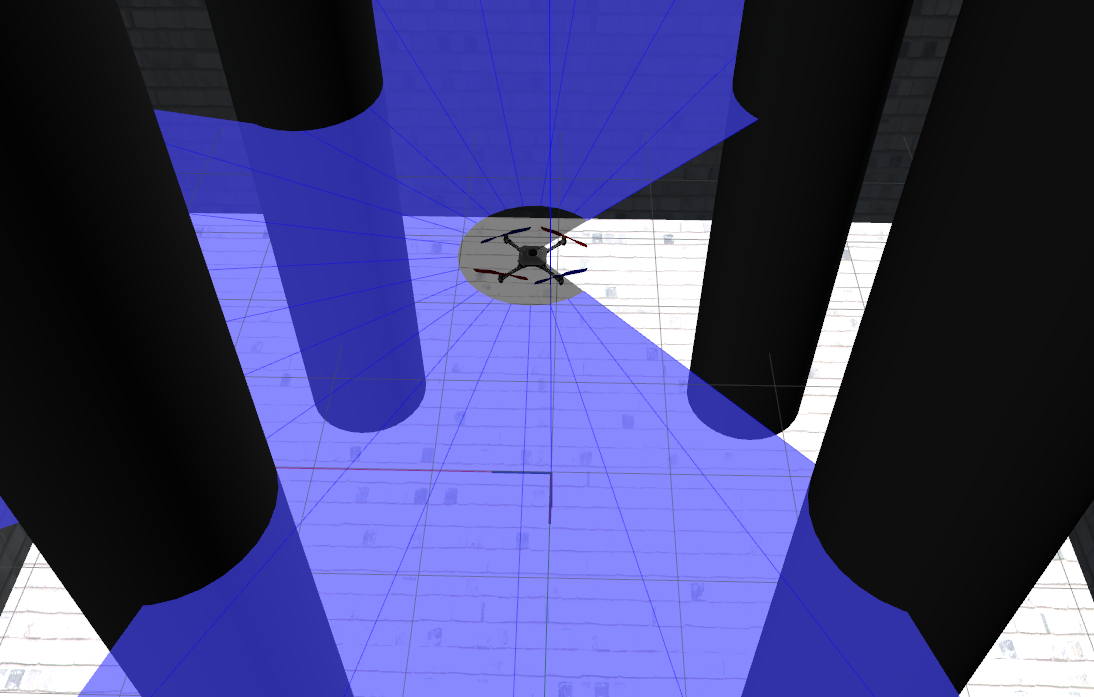}
	\end{minipage}}
  \hfill 
  \subfloat[Second environment from above.\label{fig:env2_upper}]{
	\begin{minipage}[c][0.65\width]{
	   0.5\textwidth}
	   \centering
	   \includegraphics[width=0.62\textwidth]{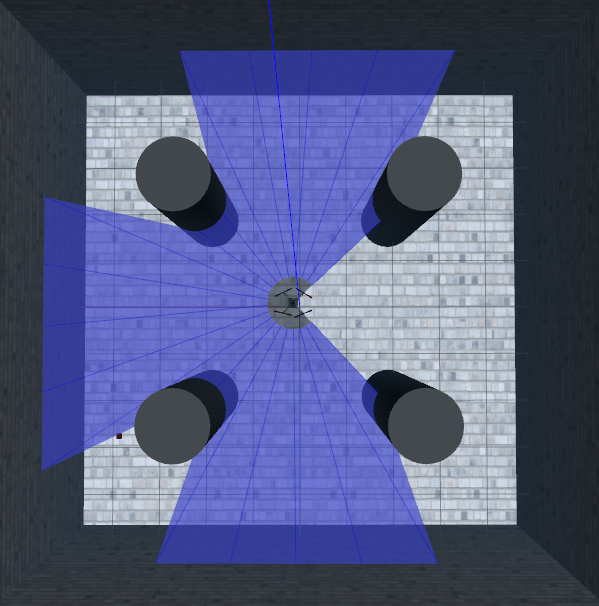}
	\end{minipage}}
 \hfill
\caption{3D Simulated Environments.}
\label{fig:envs}
\end{figure}

Two environments were used to train and test our agents. The first environment is essentially a box of 10 meters of side, with free space to perform the navigation. The second environment is just like the first environment but with 4 fixed obstacles added to it. These obstacles were set to represent drilling risers of an industrial scenario, where this approach could be applied. 
Both environments have $5 \times 5m$ of area, as can be seen in Figure \ref{fig:envs}. The blue lines are the vehicle's sensor rays.

We also used simulated wind to increase the difficulty. The wind was simulated using the Ornstein-Uhlenbeck noise in the three axis, with a velocity set to range from $-0.175$ to $0.175$ m/s. This is considerably high for an indoor environment, representing $70$\% of the maximum velocity that the vehicle can achieve. Empirically, much higher values made the scenario almost unnavigable, while lower values present no much influence on the overall results.

\section{Results}
\label{results}

In this section, we address the results obtained during the evaluation phase. The evaluation was carried out with two tasks and several statistics were recorded for each one. In the first task, the navigation task, the vehicle should navigate from a point in space to another, while multiples points should be visited in the second task, the waypoint navigation task. The simulated wind was used in both task during the test and the evaluation. Also, an agent was trained for each one of the scenarios. 

Both tasks were evaluated in both scenarios for a total of 100 trials each. The total of successful trails represents the percentage of navigation where the vehicle reached the desired goal in the first task, while in the second task the success represents the percentage of the trail where the trajectory was completed, from the first point through the final one. The average time represents the total time that was needed to complete all the 100 trials divided by 100. For the second task only, it was also registered the average percentage of the path navigated. For example, if a percentage of 10\%, the vehicle has navigated on average through 10\% of all points. This metric was used to evaluate the learning approaches as well as the BUG2 algorithm. We defined an initial position for the vehicle at the point (0.0, 0.0, 2.5) and placed the front part of it pointed towards the positive $x$-axis.  

\begin{figure}[h]
\vspace{-5mm}
  \subfloat[First environment.\label{fig:reward_1}]{
	\begin{minipage}[c][0.72\width]{
	   0.48\textwidth}
	   \centering
	   \includegraphics[width=1.0\textwidth]{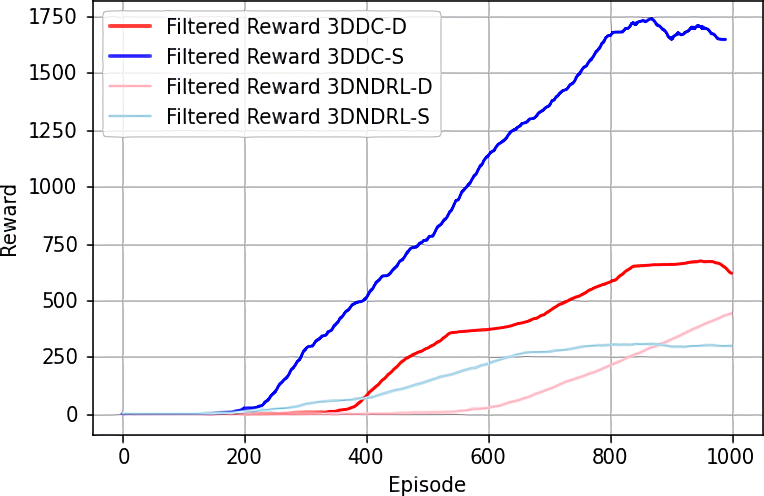}
	\end{minipage}}
 \hfill 	
  \subfloat[Second environment.\label{fig:reward_2}]{
	\begin{minipage}[c][0.72\width]{
	   0.48\textwidth}
	   \centering
	   \includegraphics[width=1.0\textwidth]{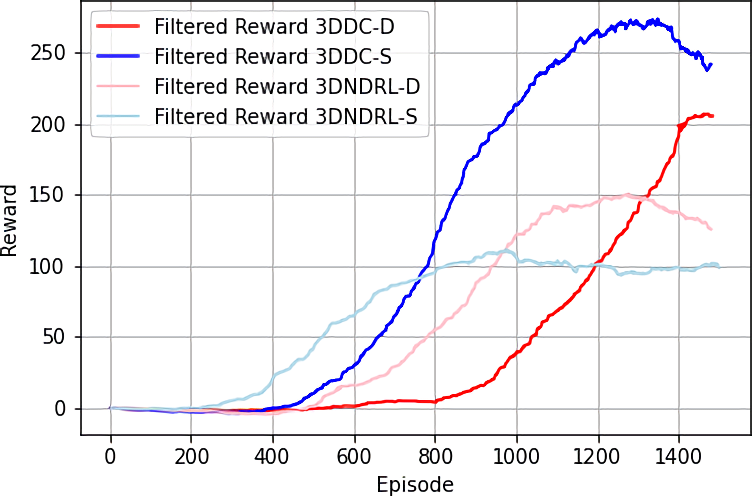}
	\end{minipage}}
\caption{Moving average of the reward over 300 episodes of the training.}
\label{fig:rewards}
\end{figure}

During the training phase, the data from the rewards was collected. We collected the data for 1000 episodes in the first scenario and 1500 episodes in the second scenario. The final episode of training was chosen based on the stagnation of the average reward received. 
Figures \ref{fig:reward_1} and \ref{fig:reward_2} show the rewards' moving average of the previous 300 episodes for the first and second scenario, respectively. 

It is possible to observe in the Figures \ref{fig:reward_1} and \ref{fig:reward_2} that the 3DDC-S and 3DDC-D approaches received the overall higher rewards, manly our newly proposed approaches based on an RNN. As the obtained reward is intrinsically attached to the performance of the agent, we are able to estimate the degree of learning of the agent in the environment. For that, we show the statistics of the navigation task (Table \ref{table:nav_air}) and for the second task (Table \ref{table:multinav_air_3D}). 

\begin{table}[!h]
\centering
\caption{Statistics of the performed navigation for the first task.}
\label{table:nav_air}
\begin{tabular}{c c c c} 
\toprule
Scenario & Approach & Success & Average Time (s) \\
\midrule
1 & 3DDC-S & 100\% & $56.48 \pm 25.71$ \\
1 & 3DDC-D & 100\% & $36.32  \pm 18.67$ \\
1 & 3DNDRL-S & 100\% & $27.03 \pm 18.67$ \\
1 & \textbf{3DNDRL-D} & 100\% & $\textbf{13.69} \pm \textbf{0.85}$ \\
1 & 3DBUG2 & 100\% & $21.92 \pm 1.341$ \\
2 & \textbf{3DDC-S} & \textbf{69\% }& $53.82 \pm 25.10$ \\
2 & \textbf{3DDC-D} & \textbf{63\%} & $\textbf{19.56} \pm \textbf{4.28}$ \\
2 & 3DNDRL-S & 19\% & $39.43 \pm 22.21$ \\
2 & 3DNDRL-D & 17\% & $49.27 \pm 38.32$ \\
2 & 3DBUG2 & 32\% & $51.93 \pm 26.83$ \\
\bottomrule
\end{tabular}
\end{table}

\begin{table}[!htb]
\centering
\caption{Statistics of the performed navigation for the second task.}
\label{table:multinav_air_3D}
\begin{tabular}{c c c c c} 
\toprule
Scenario & Approach & Success & Average Time (s) & Percentage \\
\midrule
1 & 3DDC-S  & 94\% & $155.82 \pm 60.69$ & 95.34\% \\
1 & 3DDC-D  & 27\% & $200.44 \pm 17.51$ & 71.42\% \\
1 & 3DNDRL-S  & 10\% & $135.22 \pm 61.15$ & 47.08\% \\
1 & \textbf{3DNDRL-D} & 77\% & $\textbf{76.99} \pm \textbf{29.23}$ & 86.44\% \\
1 & \textbf{3DBUG2}  & \textbf{100\%} & $145.94 \pm 19.45$ & \textbf{100\%} \\
2 & \textbf{3DDC-S}  & \textbf{26\%} & $145.16 \pm 92.29$ & \textbf{37.95\% }\\
2 & \textbf{3DDC-D}  & \textbf{41\%} & $\textbf{85.88} \pm \textbf{67.88}$ & \textbf{51.632\%} \\
2 & 3DNDRL-S  & 0\% & - & 3.46\% \\ 
2 & 3DNDRL-D & 0\% & - & 2.24\% \\ 
2 & 3DBUG2  & 3\% & $90.95 \pm 93.07$ & 11.42\% \\
\bottomrule
\end{tabular}
\end{table}

In the first task, we can see that the Deep-RL approaches and the BUG2 algorithm performed with 100\% success in the first scenario, with distinctiveness for the DDPG-based agent which yields the lowest average time and standard deviation. In the second environment, however, it becomes more clear how better are the agents based on the RNN are. Besides the average 60\% plus of success, the Deep-RL-LSTM-based approaches outperformed by almost double the BUG2 algorithm, with a special highlight for the 3DDC-D approach that performed the task consistently faster than the other approaches.

For the second task, it becomes even more clear the advantages of a Deep-RL approach with RNN architecture. We can see especially in the second environment how better the agents performed, even when compared with the BUG2 algorithm. The 3DDC-D approach managed to complete on average 51.65\% of the trajectory while the 3DDC-S agent navigates almost 40\%. Both results are considerably higher than the DDPG(8.36\%) and BUG2(11.43\%), outperforming by almost a factor of five. We can also see in table \ref{table:nav_air} and \ref{table:multinav_air_3D} that the average time of our adapted 3DNDRL-D was better in the first environment, while the 3DDC-D presented the fastest results in the second. That can be due to fact that a simpler deterministic approach tends to perform better in a simple environment, while a most robust one can be more efficient in a more complex one. It is also interesting to notice that the stochastic approach was considerably slower given its characteristics, but showed to be more trusty than the deterministic one. Overall, it is important to highlight how well agents managed to understand the vehicle and the difficulties that the environment provided. We can see based on the results of the BUG2 algorithm, how challenging was to perform this task in an environment with real simulated wind and dynamics. 

\begin{figure*}[p]
  \subfloat[Trajectory 3DNDRL-S method. \label{fig:nav_position_sac_stage_1_air3D_tanh_3layers}]{
	\begin{minipage}[c][0.7\width]{
	   0.47\textwidth}
	   \centering
	   \includegraphics[width=1.0\textwidth]{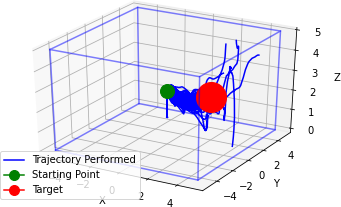}
	\end{minipage}}
 \hfill	
 \subfloat[Upper view trajectory 3DNDRL-S method. \label{fig:navupper_position_sac_stage_1_air3D_tanh_3layers}]{
	\begin{minipage}[c][0.7\width]{
	   0.47\textwidth}
	   \centering
	   \includegraphics[width=1.0\textwidth]{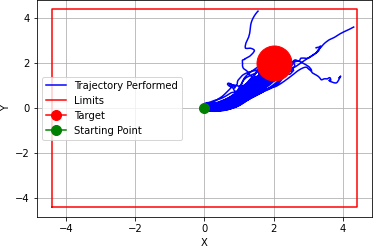}
	\end{minipage}}
 \hfill
 \vspace{-4mm}
 \subfloat[Trajectory 3DNDRL-D method. \label{fig:nav_position_ddpg_stage_1_air3D_tanh_3layers}]{
	\begin{minipage}[c][0.7\width]{
	   0.49\textwidth}
	   \centering
	   \includegraphics[width=1.0\textwidth]{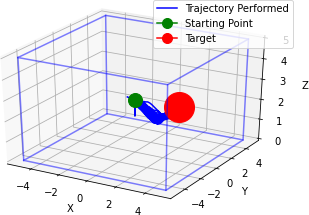}
	\end{minipage}}
 \hfill	
 \subfloat[Upper view trajectory 3DNDRL-D method. \label{fig:navupper_position_ddpg_stage_1_air3D_tanh_3layers}]{
	\begin{minipage}[c][0.7\width]{
	   0.49\textwidth}
	   \centering
	   \includegraphics[width=1.0\textwidth]{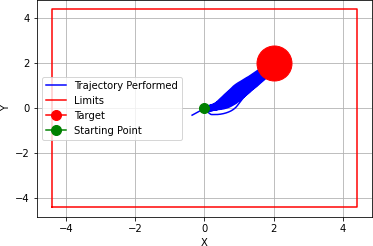}
	\end{minipage}}
 \hfill
  \vspace{-4mm}
\subfloat[Trajectory 3DDC-D method. \label{fig:nav_position_td3_stage_1_air3D_tanh_lstm}]{
	\begin{minipage}[c][0.7\width]{
	   0.47\textwidth}
	   \centering
	   \includegraphics[width=1.0\textwidth]{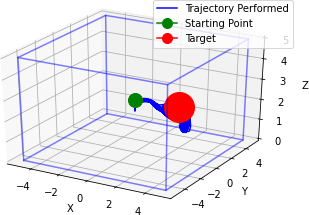}
	\end{minipage}}
 \hfill 
 \subfloat[Upper view trajectory 3DDC-D method. \label{fig:navupper_position_td3_stage_1_air3D_tanh_lstm}]{
	\begin{minipage}[c][0.7\width]{
	   0.47\textwidth}
	   \centering
	   \includegraphics[width=1.0\textwidth]{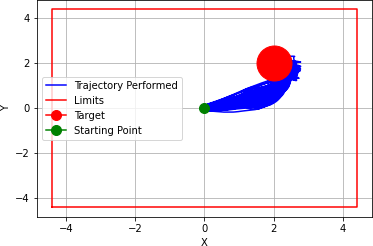}
	\end{minipage}}
 \hfill 
  \vspace{-1mm}
  \subfloat[Trajectory 3DDC-S method. \label{fig:nav_position_sac_stage_1_air3D_tanh_lstm}]{
	\begin{minipage}[c][0.7\width]{
	   0.47\textwidth}
	   \centering
	   \includegraphics[width=1.0\textwidth]{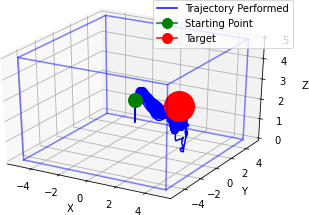}
	\end{minipage}}
 \hfill	
 \subfloat[Upper view trajectory 3DDC-S method. \label{fig:navupper_position_sac_stage_1_air3D_tanh_lstm}]{
	\begin{minipage}[c][0.7\width]{
	   0.47\textwidth}
	   \centering
	   \includegraphics[width=1.0\textwidth]{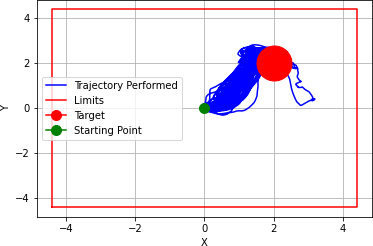}
	\end{minipage}}
 \hfill
  \vspace{-1mm}
\subfloat[Trajectory BUG2 method. \label{fig:nav_position_bug_stage_1_air3D}]{
\begin{minipage}[c][0.7\width]{
   0.47\textwidth}
   \centering
   \includegraphics[width=1.0\textwidth]{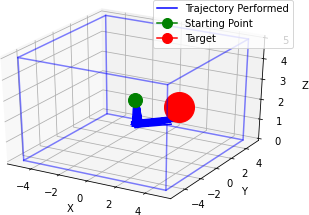}
\end{minipage}}
 \hfill
\subfloat[Upper view trajectory BUG2 method. \label{fig:navupper_position_bug_stage_1_air3D}]{
	\begin{minipage}[c][0.7\width]{
	   0.47\textwidth}
	   \centering
	   \includegraphics[width=1.0\textwidth]{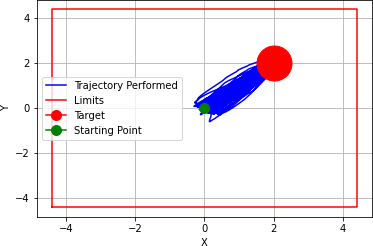}
	\end{minipage}}
\caption{Navigation task in the first environment.}
\label{fig:traj_nav_1}
\end{figure*}

\begin{figure*}[p]
  \subfloat[Trajectory 3DNDRL-S method. \label{fig:nav_position_sac_stage_2_air3D_tanh_3layers}]{
	\begin{minipage}[c][0.7\width]{
	   0.47\textwidth}
	   \centering
	   \includegraphics[width=1.0\textwidth]{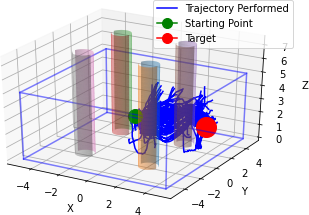}
	\end{minipage}}
 \hfill	
 \subfloat[Upper view trajectory 3DNDRL-S method. \label{fig:navupper_position_sac_stage_2_air3D_tanh_3layers}]{
	\begin{minipage}[c][0.7\width]{
	   0.47\textwidth}
	   \centering
	   \includegraphics[width=1.0\textwidth]{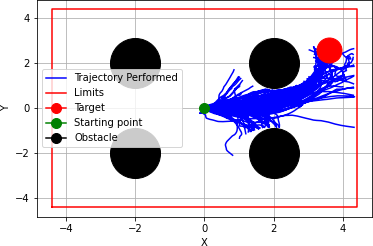}
	\end{minipage}}
 \hfill
 \vspace{-4mm}
 \subfloat[Trajectory 3DNDRL-D method. \label{fig:nav_position_ddpg_stage_2_air3D_tanh_3layers}]{
	\begin{minipage}[c][0.7\width]{
	   0.49\textwidth}
	   \centering
	   \includegraphics[width=1.0\textwidth]{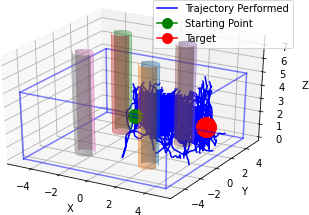}
	\end{minipage}}
 \hfill	
 \subfloat[Upper view trajectory 3DNDRL-D method. \label{fig:navupper_position_ddpg_stage_2_air3D_tanh_3layers}]{
	\begin{minipage}[c][0.7\width]{
	   0.49\textwidth}
	   \centering
	   \includegraphics[width=1.0\textwidth]{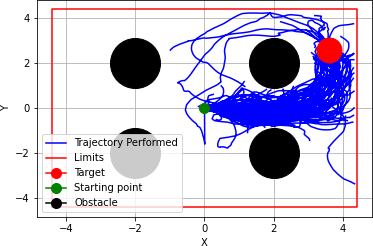}
	\end{minipage}}
 \hfill
  \vspace{-4mm}
\subfloat[Trajectory 3DDC-D method. \label{fig:nav_position_td3_stage_2_air3D_tanh_lstm}]{
	\begin{minipage}[c][0.7\width]{
	   0.47\textwidth}
	   \centering
	   \includegraphics[width=1.0\textwidth]{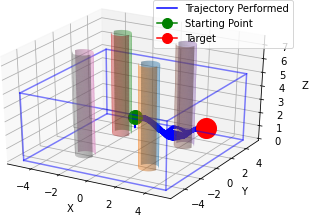}
	\end{minipage}}
 \hfill 
 \subfloat[Upper view trajectory 3DDC-D method. \label{fig:navupper_position_td3_stage_2_air3D_tanh_lstm}]{
	\begin{minipage}[c][0.7\width]{
	   0.47\textwidth}
	   \centering
	   \includegraphics[width=1.0\textwidth]{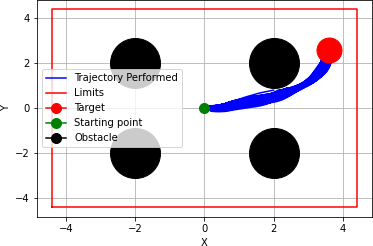}
	\end{minipage}}
 \hfill 
  \vspace{-2mm}
  \subfloat[Trajectory 3DDC-S method. \label{fig:nav_position_sac_stage_2_air3D_tanh_lstm}]{
	\begin{minipage}[c][0.7\width]{
	   0.47\textwidth}
	   \centering
	   \includegraphics[width=1.0\textwidth]{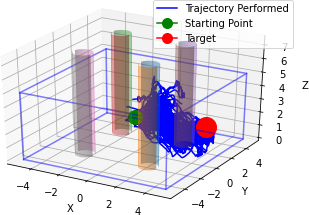}
	\end{minipage}}
 \hfill	
 \subfloat[Upper view trajectory 3DDC-S method. \label{fig:navupper_position_sac_stage_2_air3D_tanh_lstm}]{
	\begin{minipage}[c][0.7\width]{
	   0.47\textwidth}
	   \centering
	   \includegraphics[width=1.0\textwidth]{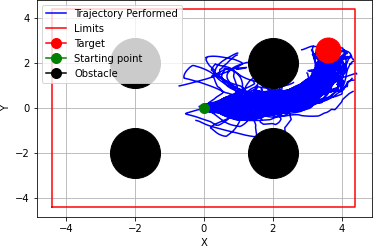}
	\end{minipage}}
 \hfill
  \vspace{-2mm}
   \subfloat[Trajectory BUG2 method. \label{fig:nav_position_bug_stage_2_3d}]{
	\begin{minipage}[c][0.7\width]{
	   0.47\textwidth}
	   \centering
	   \includegraphics[width=1.0\textwidth]{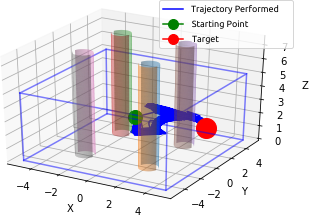}
	\end{minipage}}
 \hfill
\subfloat[Upper view trajectory BUG2 method. \label{fig:navupper_position_bug_stage_2_3d}]{
	\begin{minipage}[c][0.7\width]{
	   0.47\textwidth}
	   \centering
	   \includegraphics[width=1.0\textwidth]{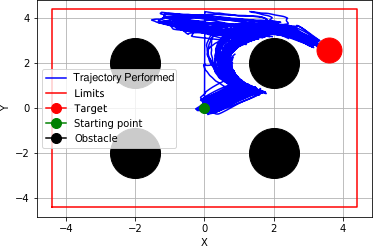}
	\end{minipage}}
	
\caption{Navigation task in the second environment.}
\label{fig:traj_nav_2}
\end{figure*}

\begin{figure*}[p]
  \subfloat[Trajectory 3DNDRL-S method. \label{fig:multinav_position_sac_stage_1_air3D_tanh_3layers}]{
	\begin{minipage}[c][0.65\width]{
	   0.47\textwidth}
	   \centering
	   \includegraphics[width=1.0\textwidth]{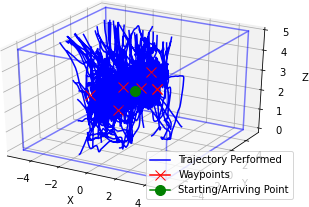}
	\end{minipage}}
 \hfill	
 \subfloat[Upper view trajectory 3DNDRL-S method. \label{fig:multinavupper_position_sac_stage_1_air3D_tanh_3layers}]{
	\begin{minipage}[c][0.65\width]{
	   0.47\textwidth}
	   \centering
	   \includegraphics[width=1.0\textwidth]{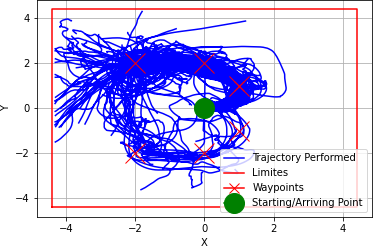}
	\end{minipage}}
 \hfill
 \vspace{-3mm}
 \subfloat[Trajectory 3DNDRL-D method. \label{fig:multinav_position_ddpg_stage_1_air3D_tanh_3layers}]{
	\begin{minipage}[c][0.65\width]{
	   0.49\textwidth}
	   \centering
	   \includegraphics[width=1.0\textwidth]{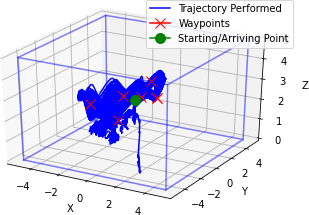}
	\end{minipage}}
 \hfill	
 \subfloat[Upper view trajectory 3DNDRL-D method. \label{fig:multinavupper_position_ddpg_stage_1_air3D_tanh_3layers}]{
	\begin{minipage}[c][0.65\width]{
	   0.49\textwidth}
	   \centering
	   \includegraphics[width=1.0\textwidth]{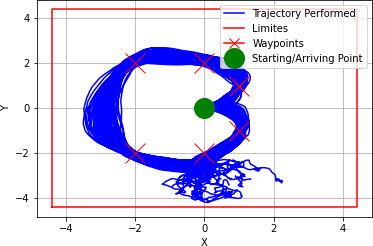}
	\end{minipage}}
 \hfill
 \vspace{-3mm}
\subfloat[Trajectory 3DDC-D method. \label{fig:multinav_position_td3_stage_1_air3D_tanh_lstm}]{
	\begin{minipage}[c][0.65\width]{
	   0.47\textwidth}
	   \centering
	   \includegraphics[width=1.0\textwidth]{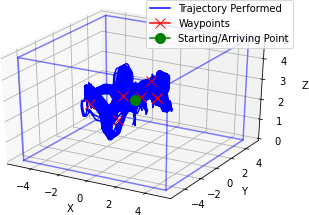}
	\end{minipage}}
 \hfill 
 \subfloat[Upper view trajectory 3DDC-D method. \label{fig:multinavupper_position_td3_stage_1_air3D_tanh_lstm}]{
	\begin{minipage}[c][0.65\width]{
	   0.47\textwidth}
	   \centering
	   \includegraphics[width=1.0\textwidth]{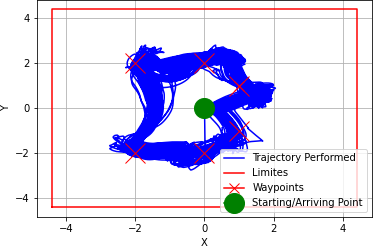}
	\end{minipage}}
 \hfill
  \subfloat[Trajectory 3DDC-S method. \label{fig:multinav_position_sac_stage_1_air3D_tanh_lstm}]{
	\begin{minipage}[c][0.65\width]{
	   0.47\textwidth}
	   \centering
	   \includegraphics[width=1.0\textwidth]{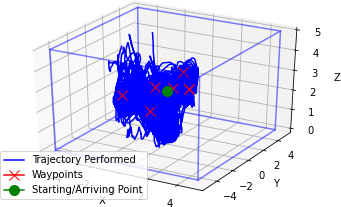}
	\end{minipage}}
 \hfill	
 \subfloat[Upper view trajectory 3DDC-S method. \label{fig:multinavupper_position_sac_stage_1_air3D_tanh_lstm}]{
	\begin{minipage}[c][0.65\width]{
	   0.47\textwidth}
	   \centering
	   \includegraphics[width=1.0\textwidth]{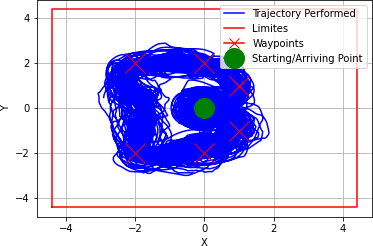}
	\end{minipage}}
 \hfill
\subfloat[Trajectory BUG2 method. \label{fig:multinav_position_bug_stage_1_air3D}]{
	\begin{minipage}[c][0.65\width]{
	   0.47\textwidth}
	   \centering
	   \includegraphics[width=1.0\textwidth]{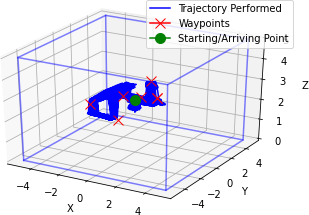}
	\end{minipage}}
 \hfill
\subfloat[Upper view trajectory BUG2 method. \label{fig:multinavupper_position_bug_stage_1_air3D}]{
	\begin{minipage}[c][0.65\width]{
	   0.47\textwidth}
	   \centering
	   \includegraphics[width=1.0\textwidth]{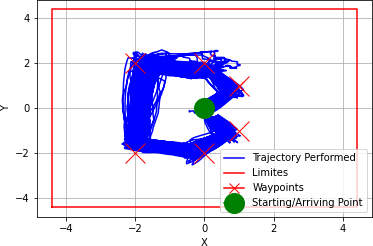}
	\end{minipage}}
 \hfill
\caption{Waypoint navigation task in the first environment.}
\label{fig:multi_nav_traj_1}
\end{figure*}

\begin{figure*}[p]
  \subfloat[Trajectory 3DNDRL-S method. \label{fig:multinav_position_sac_stage_2_air3D_tanh_3layers}]{
	\begin{minipage}[c][0.7\width]{
	   0.47\textwidth}
	   \centering
	   \includegraphics[width=1.0\textwidth]{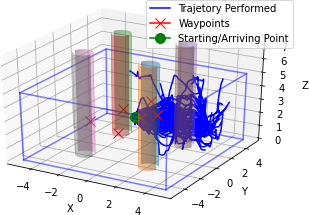}
	\end{minipage}}
 \hfill	
 \subfloat[Upper view trajectory 3DNDRL-S method. \label{fig:multinavupper_position_sac_stage_2_air3D_tanh_3layers}]{
	\begin{minipage}[c][0.7\width]{
	   0.47\textwidth}
	   \centering
	   \includegraphics[width=1.0\textwidth]{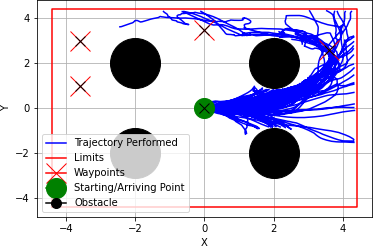}
	\end{minipage}}
 \hfill
 \vspace{-4mm}
 \subfloat[Trajectory 3DNDRL-D method. \label{fig:multinav_position_ddpg_stage_2_air3D_tanh_3layers}]{
	\begin{minipage}[c][0.7\width]{
	   0.49\textwidth}
	   \centering
	   \includegraphics[width=1.0\textwidth]{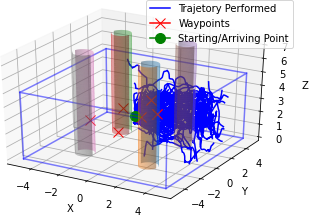}
	\end{minipage}}
 \hfill	
 \subfloat[Upper view trajectory 3DNDRL-D method. \label{fig:multinavupper_position_ddpg_stage_2_air3D_tanh_3layers}]{
	\begin{minipage}[c][0.7\width]{
	   0.49\textwidth}
	   \centering
	   \includegraphics[width=1.0\textwidth]{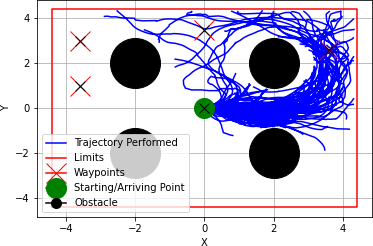}
	\end{minipage}}
 \hfill
 \vspace{-4mm}
\subfloat[Trajectory 3DDC-D method. \label{fig:multinav_position_td3_stage_2_air3D_tanh_lstm}]{
	\begin{minipage}[c][0.7\width]{
	   0.47\textwidth}
	   \centering
	   \includegraphics[width=1.0\textwidth]{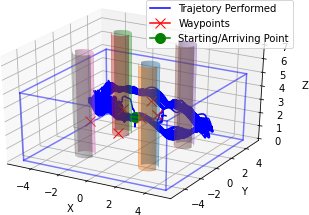}
	\end{minipage}}
 \hfill 
 \subfloat[Upper view trajectory 3DDC-D method. \label{fig:multinavupper_position_td3_stage_2_air3D_tanh_lstm}]{
	\begin{minipage}[c][0.7\width]{
	   0.47\textwidth}
	   \centering
	   \includegraphics[width=1.0\textwidth]{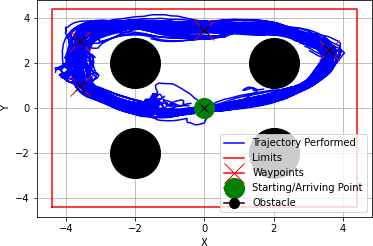}
	\end{minipage}}
 \hfill 
  \vspace{-2mm}
  \subfloat[Trajectory 3DDC-S method. \label{fig:multinav_position_sac_stage_2_air3D_tanh_lstm}]{
	\begin{minipage}[c][0.7\width]{
	   0.47\textwidth}
	   \centering
	   \includegraphics[width=1.0\textwidth]{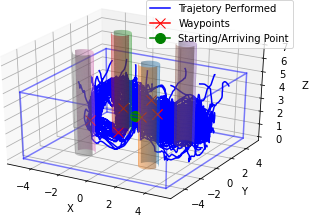}
	\end{minipage}}
 \hfill	
 \subfloat[Upper view trajectory 3DDC-S method. \label{fig:multinavupper_position_sac_stage_2_air3D_tanh_lstm}]{
	\begin{minipage}[c][0.7\width]{
	   0.47\textwidth}
	   \centering
	   \includegraphics[width=1.0\textwidth]{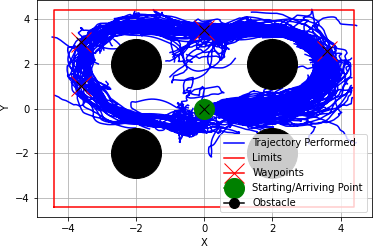}
	\end{minipage}}
 \hfill
  \vspace{-2mm}
   \subfloat[Trajectory BUG2 method. \label{fig:multinav_position_bug_stage_2_air3d}]{
	\begin{minipage}[c][0.7\width]{
	   0.47\textwidth}
	   \centering
	   \includegraphics[width=1.0\textwidth]{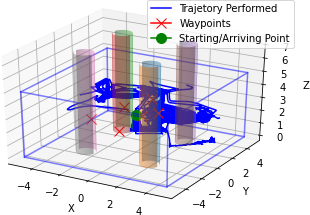}
	\end{minipage}}
 \hfill
\subfloat[Upper view trajectory BUG2 method. \label{fig:multinavupper_position_bug_stage_2_air3d}]{
	\begin{minipage}[c][0.7\width]{
	   0.47\textwidth}
	   \centering
	   \includegraphics[width=1.0\textwidth]{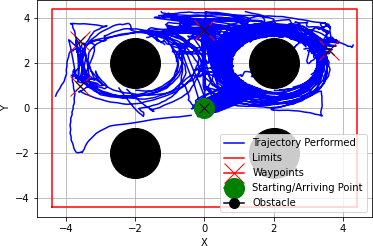}
	\end{minipage}}
 \hfill
\caption{Waypoint navigation task in the second environment.}
\label{fig:multi_nav_traj_2}
\end{figure*}

Finally we can observe both task performed in both environments in the Figures \ref{fig:traj_nav_1}, \ref{fig:traj_nav_2}, \ref{fig:multi_nav_traj_1} and \ref{fig:multi_nav_traj_2}. It shows the 100 trials performed for each one of the approaches. We can see even better the more aleatory characteristic of the stochastic approach when compared to the deterministic one. The 100 samples for both tasks show that this first approach presents a more exploratory trajectory, which affects the overall time to perform it. It is also possible to verify that even the deterministic approach and the BUG2 algorithm suffered considerably the influence of the simulated wind, showing trajectories with consistent changes.

In the second task specifically, we can see in Figure \ref{fig:nav_position_td3_stage_2_air3D_tanh_lstm} one of the main contribution of this work. It is possible to outline the ability to learn the simulated wind and to adapt to perform the navigation correctly. When compared to the the stochastic approach (Figure \ref{fig:nav_position_sac_stage_2_air3D_tanh_lstm}) or even to the BUG2 algorithm (\ref{fig:nav_position_bug_stage_2_3d}) we can see the more stable trajectories which provided the overall better results. For the second task in the second environment the same can be said, while 3DDC-D and 3DDC-S learned the task and the environmental difficulties, the previous methods and the traditional method presented consistently more difficulties.

The trials of the 3DDC-D approach (Figures \ref{fig:multinav_position_td3_stage_2_air3D_tanh_lstm} \ref{fig:multinavupper_position_td3_stage_2_air3D_tanh_lstm}) and the 3DDC-S approach (Figures \ref{fig:multinav_position_sac_stage_2_air3D_tanh_lstm} \ref{fig:multinavupper_position_sac_stage_2_air3D_tanh_lstm}) in the second scenario shows how effective these two approaches became when comparing with the others. Both managed to understand the task, the environment and to avoid obstacles with great distinctiveness.

\section{Conclusion}
\label{conclusion}

In this work, we proposed a novel methodology based on the state-of-art Deep-RL techniques to address the mapless 3D navigation and waypoint 3D navigation tasks for UAVs. We showed that approaches with double critic and RNNs are better suited than the state-of-art approaches with fully-connected ANNs, commonly used for terrestrial mobile robots. Our proposed system is able to perform navigation-related tasks by only using the vehicles' relative localization data and some range findings, without making the use of imaged-based sensing which can demand a costly hardware capability to achieve good performance. We can conclude that our approaches based on double critic actor-critic SAC and TD3 algorithms also managed to bypass obstacles and reach the desired goal. 

Specifically, it is also possible to conclude that our novel approaches outperformed a DDPG-based approach and the BUG2 algorithm. The good results achieved can be due to a simple sensory-based structure and a simple rewarding system.

As future works, we are aiming to test in our real vehicle and develop a similar approach for Unmanned Underwater Vehicles (UUVs) and Hybrid Unmanned Aerial Underwater Vehicles (HUAUVS).

\section*{Aknowledgement}

The authors would like to thank to National Council for Scientific and Technological Development (CNPq), the Coordination for the Improvement of Higher Education Personnel (CAPES) - Finance Code 001, PRH-ANP and all  participants  of  VersusAI.

\clearpage
\section*{Declarations}

\textbf{Ethical Approval} ~The article has the approval of all the authors.\\

\noindent
\textbf{Consent to Participate} ~All the authors gave their consent to participate in this article.\\

\noindent
\textbf{Consent to Publish} ~The authors gave their authorization for the publishing of this article.\\

\noindent
\textbf{Authors’ contributions:}

\begin{itemize}
    \item \noindent
    \textbf{Ricardo Bedin Grando} ~conceived the research, writing the article, designed and program the experiments, collected and processed the test data.
    \item \noindent
    \textbf{Junior Costa de Jesus} ~writing of the article, collected and processed the test data.
    \item \noindent
    \textbf{Victor Augusto Kich} ~write the article, program the experiments, collected and processed the test data.
    \item \noindent
    \textbf{Alisson Henrique Kolling} ~write the article, program the experiments, collected and processed the test data.
    \item \noindent
    \textbf{Paulo Lilles Jorge Drews Jr.} ~conceived the research, writing of the article and discussion of the main ideas of the article.
\end{itemize}

\noindent
\textbf{Funding} ~National Council for Scientific and Technological Development (CNPq), Coordination for the Improvement of Higher Education Personnel (CAPES) and National Agency of Petroleum, Natural Gas and Biofuels (PRH-ANP).\\

\noindent
\textbf{Competing interest} ~There are not conflict of interest or competing interest.\\

\noindent
\textbf{Available of data and material} ~available in GitHub \url{https://github.com/ricardoGrando/hydrone\_deep\_rl\_jint}.



\bibliographystyle{IEEEtran}
\bibliography{IEEEabrv,bibliography}

\end{document}